# Dire n'est pas concevoir [1]

Christophe Roche[1]

[1] Equipe Condillac, Université de Savoie, Listic
73 376 Le Bourget du Lac cedex
`roche@univ-savoie.fr`
`http://www.ontology.univ-savoie.fr`

**Résumé** : Nous verrons dans le cadre de cet article que l'extraction de connaissances à partir de textes relève avant tout de la linguistique textuelle dont un des principes est l'incomplétude des textes ; et qu'il est difficile de prendre en considération les connaissances extralinguistiques nécessaires à leur compréhension. L'utilisation de figures de style, telles que la métonymie et la synecdoque, modifie profondément la perception que nous pouvons avoir des concepts du domaine. La structure conceptuelle extraite à partir de textes n'est pas une ontologie, comprise comme connaissance consensuelle et générale (non contingente) partagée par une communauté d'acteurs, mais est du ressort de la sémantique lexicale : *structure conceptuelle du domaine et structure lexicale ne se superposent pas*. Si le langage naturel n'est pas un mode d'expression adapté à la description d'ontologies, nous verrons également que les langages formels posent un certain nombre de problèmes qui leur sont propres. La modélisation d'ontologies reste avant tout un problème épistémologique.

**Mots-clés** : Ingénierie des connaissances, Ontologies, Terminologies, Construction d'ontologies à partir de textes, Linguistique textuelle, Langages formels, Ontologies par différenciation spécifique.

## 1 Introduction

> « *Contentons-nous de convenir que ce n'est pas des noms qu'il faut partir, mais qu'il faut et apprendre et rechercher les choses en partant d'elles-mêmes bien plutôt que des noms.* ». Cratyle, 439, b.

La construction d'ontologies est une tâche difficile pouvant être longue et coûteuse. C'est la raison pour laquelle il existe un certain nombre de travaux qui visent, si ce n'est à automatiser, du moins à faciliter cette tâche. Si l'on considère que les documents scientifiques et techniques d'un domaine en véhiculent les connaissances, l'extraction d'ontologies à partir de textes semble être une idée des plus prometteuses. L'idée principale est que les termes dénotent des concepts, et que les relations linguistiques entre termes traduisent une relation entre concepts. La figure ci-dessous décrit ce processus de rétro ingénierie de connaissances à partir de textes.

---

[1] Cet article reprend, en français et en tenant compte des remarques et des débats qui ont suivi, les thèmes développés dans deux articles, « Lexical structure and conceptual structure » présenté lors de la session spéciale "Ontology and Text" de la XIX[ème] Conférence IEA/AIE à Annecy 27-30 June 2006 et « How words map concepts » présenté à VORTE 2006 - 10th IEEE EDOC Conference. Hong Kong. 16-20 October 2006.

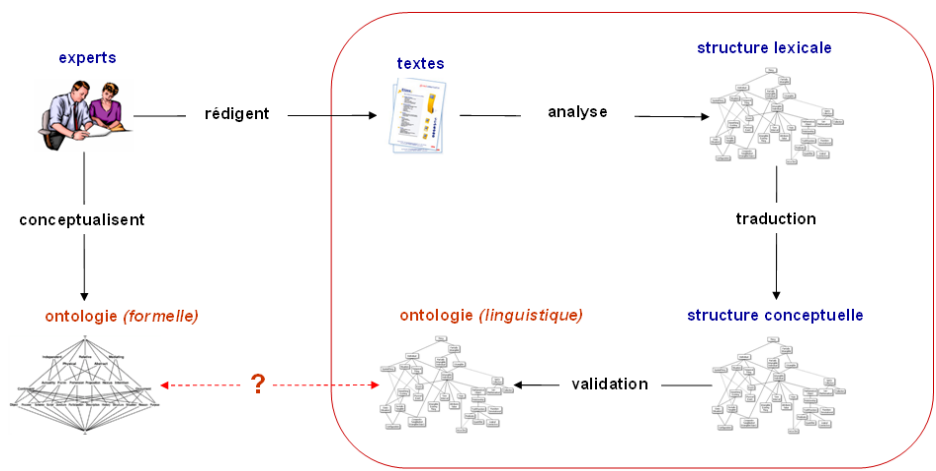

**Fig. 1** – Processus de construction d'ontologies à partir de textes.

Un tel processus soulève néanmoins un certain nombre de questions. La principale est de savoir dans quelle mesure une ontologie construite à partir de textes et une ontologie définie directement par les experts sont comparables. Autrement dit : Quelles sont les conséquences de l'utilisation d'un langage donné, qu'il soit naturel ou formel ? Que perdons-nous (et introduisons-nous) lors de l'écriture de textes et en quoi les langages formels conditionnent la conceptualisation ? Autant de questions auxquelles nous essaierons d'apporter des éléments de réponse.

## 2   La construction d'ontologies à partir de textes

Ce chapitre rappelle brièvement les principales étapes du processus d'acquisition d'ontologies à partir de textes et les principes sur lesquels elles reposent.

### 2.1   Le contexte industriel

Afin d'illustrer nos propos, nous nous appuierons sur une application industrielle réalisée par la société Ontologos corp.[2] pour EDF Recherche & Développement (Dourgnon-Hanoune *et al.* 2006), (Dourgnon-Hanoune *et al.* 2005). Le principal objectif du projet est la réappropriation d'une partie des connaissances ontologiques et terminologiques du domaine du contrôle commande. Le problème est d'autant plus difficile que cette connaissance n'est pas directement accessible sous une forme exploitable, mais diffuse en particulier à travers différents types de documents. Dans le cadre de cet article nous nous intéresserons principalement au corpus[3] des relais électromagnétiques. L'ontologie construite[4] vise en particulier deux objectifs : une explicitation des concepts décrivant les différents types de relais qui soit partageable et consensuelle pour différentes communautés de pratiques (ingénieurs, techniciens, opérateur, fournisseurs) et une gestion ontologique des documents du corpus.

---

[2] http://www.ontologos-corp.com

**2.2  La structure lexicale**

La première étape a pour principal objectif d'identifier les termes[5] pouvant dénoter des concepts[6] et les relations linguistiques entre ces termes.

L'extraction de candidats termes et de relations linguistiques à partir de corpus est un domaine de recherche des plus actifs (Aussenac-Gilles & Soergel 2005), (Buitelaar 2005), (Daille 2004). Elle repose sur des méthodes statistiques – exploitant par exemple une analyse distributionnelle à la Harris (Harris 1968) – et/ou linguistiques. L'utilisation d'expressions régulières est un procédé simple qui permet, après lemmatisation des documents, l'extraction de syntagmes nominaux. Des patrons lexicaux-syntaxiques « substantif adjectif », « substantif préposition substantif », etc. nous tirons les expressions « relais électromagnétique », « relais de tension », « relais à seuil », « relais tout ou rien », etc. Le résultat, qui doit être validé par les experts, est un lexique de termes de la terminologie et de mots d'usage de la langue de spécialité[7].

Ces mots d'usage se structurent en un réseau selon différentes relations linguistiques et, plus particulièrement pour notre sujet, selon les relations d'hyperonymie, de méronymie, de synonymie, etc. Comme précédemment, ces relations peuvent être *automatiquement* extraites du corpus à l'aide de méthodes syntaxiques – « un relais de tension *est* un relais » –, lexicales – traitement à l'aide d'expressions régulières des syntagmes nominaux – et/ou statistiques. Par exemple la prise en compte des expressions avec *même tête* permet de dire que « relais électromagnétique », « relais de tension », « relais tout ou rien », « relais à seuil » sont autant d'hyponymes de « relais » (fig.2).

---

[3] C'est-à-dire un ensemble de textes sélectionnés selon certains critères : appartenance à une communauté de pratique, sujet traité, genre textuel, forme, etc. (http://ahds.ac.uk/linguistic-corpora/)

[4] La construction de l'ontologie s'est faite à l'aide des outils d'Ontologos corp. LCW (Linguistic Craft Workbench) pour l'extraction de candidats termes, SNCW (Semantic Network Craft Workbench) pour la construction des réseaux lexicaux et conceptuels (à base de schémas), OCW (Ontology Craft Workbench) pour la définition d'ontologies formelles (par différenciation spécifique) et TCW (Terminology Craft Workbench) pour la construction de terminologies.

[5] Elément d'une terminologie ou lexie d'une langue de spécialité, la notion de *terme* traduit l'existence de mots liés à une pratique dont la signification est relativement stable.

[6] Dans le cadre de cet article, nous adopterons les conventions typographiques suivantes : Les lexies sont notées entre guillemets et les concepts entre les symboles inférieur et supérieur. Par exemple, le syntagme nominal « relais de tension » désigne le concept <relais de tension>.

[7] En toute rigueur, si la plupart des termes (désignations) d'une terminologie sont des mots d'usage de la langue de spécialité (LSP), la réciproque n'est pas nécessairement vraie. La construction d'ontologies à partir de textes devrait tenir compte de cette distinction.

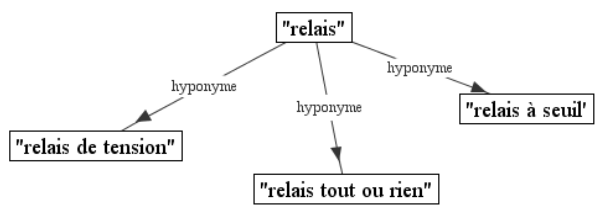

**Fig. 2** – Structure lexicale.

### 2.3 La structure conceptuelle et l'ontologie

La deuxième étape consiste à construire la structure conceptuelle à partir de la structure lexicale. Si nous considérons qu'un terme[8] dénote un concept et que la relation d'hyperonymie est une traduction linguistique[9] de la relation de subsomption, nous pouvons alors considérer que le syntagme « relais de tension » désigne le concept <relais de tension> subsumé par le concept <relais>, lui même dénoté par le terme « relais » hyperonyme de « relais de tension ». La structure conceptuelle se calque sur la structure lexicale (fig. 3).

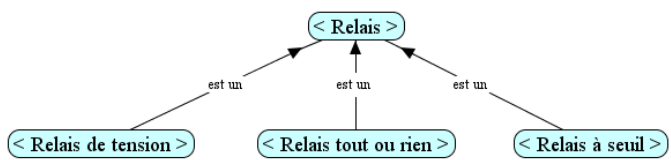

**Fig. 3** – Structure conceptuelle validée comme ontologie.

La dernière étape est la validation par les experts de la structure conceptuelle afin de l'ériger en ontologie du domaine. Durant cette étape importante, la structure conceptuelle peut être modifiée et complétée autant que nécessaire. Le cas échéant les noms des concepts sont normalisés, des variations terminologiques introduites, et des mots d'usage associés. Dans le cadre de notre application, la structure conceptuelle précédemment construite est ainsi une ontologie valide du domaine.

## 3 Est-ce si simple ?

Le processus que nous venons de décrire est clair, bien défini et de nombreux travaux sont menés sur ces différentes phases. Mais est-ce si simple ? (ce qui ne veut pas dire que les méthodes et techniques mises en œuvre lors de ces différentes phases le soient, bien au contraire). L'ontologie ainsi construite satisfait-elle vraiment nos

---

[8] Un terme peut être considéré comme un concept lexicalisé, démarche que nous pouvons rapprocher de celle de Wordnet (http://wordnet.princeton.edu/) où un concept est défini comme un *synset*, c'est-à-dire un ensemble de synonymes.

[9] Ou inversement, dire que la relation de subsomption se déduit des usages, ne change rien au résultat.

*Dire n'est pas concevoir*
attentes par rapport à l'application visée, à ses propriétés de cohérence et de réutilisabilité ainsi qu'à son adéquation, sa compatibilité par rapport à une conceptualisation qui serait directement construite par les experts à l'aide d'un langage formel ?

### 3.1    La validation de l'ontologie par l'application

Il est parfois dit qu'une ontologie est définie pour un objectif donné (Staab *et al.* 2004). Et les ontologies construites à partir de textes sont *a priori* tout à fait adaptées pour la gestion de contenus et en particulier pour la gestion des documents dont elles sont issues. L'ontologie est utilisée à la fois pour l'indexation (classification) des documents et pour leur recherche (Kiryakov *et al.* 2005). Ainsi, la recherche de documents portant sur les « relais à seuil » devrait nous retourner tous les documents relatifs à ce type de relais ; c'est-à-dire les documents rattachés au concept <relais à seuil> et aux concepts qu'il subsume. Or, en l'occurrence aucune information n'est retournée concernant les relais de tension, alors que cela devrait être le cas. Bien que la structure conceptuelle ait été validée – un <relais de tension> est bien une sorte de <relais> – elle ne répond pas aux attentes des utilisateurs qui lors de leur recherche font principalement référence à leur conception scientifique de leur domaine qui ne se calque pas sur des usages linguistiques de ces concepts[10].

### 3.2    La réutilisabilité de l'ontologie

La construction d'ontologies à partir de textes est fortement dépendante d'un corpus donné, ce qui n'a rien de surprenant. Le résultat est une ontologie difficilement partageable et réutilisable. Comment dans de telles conditions définir une ontologie commune et partageable entre les différentes communautés de pratique (techniciens et fournisseurs) partageant la *même réalité* (ici les relais) mais utilisant leur propre langue de spécialité ? Et comment définir une ontologie indépendante des variations linguistiques alors que la réalité n'a pas changé ?

### 3.3    La compatibilité avec l'ontologie définie par les experts

La nécessité de définir une conceptualisation consensuelle pour les différentes communautés de pratique partageant la même réalité, associé au fait que les utilisateurs recherchent leur information en fonction d'une conceptualisation indépendante des usages linguistiques, nous a amenés à demander aux experts de construire directement une ontologie de leur domaine dans un langage formel.

Le processus de construction est différent : les experts s'attachent davantage à identifier les concepts selon leurs caractéristiques et à les différencier qu'à les désigner en langue. Ainsi, le concept <relais de tension> n'est pas un concept au *même niveau* que le concept <relais tout ou rien> ou <relais à seuil>, mais une sorte de <relais à seuil> dont la valeur seuillée est la tension. Le résultat est une ontologie partageable entre les différentes communautés de pratique, réutilisable, mais qui ne se superpose pas avec celle construite à partir du corpus.

---

[10] Ces résultats ont été confirmés à travers différentes applications de gestion ontologique de documents scientifiques et techniques, en particulier avec le GRETh (Groupement pour la recherche sur les échangeurs thermiques) et dans le cadre d'un projet européen FP6 (ASTECH).

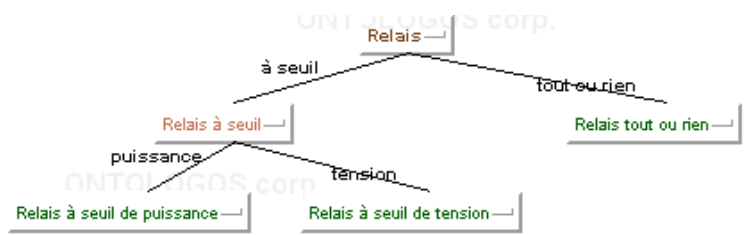

**Fig. 4** – Ontologie par différenciation spécifique.

## 4   Dire n'est pas concevoir

Le premier objectif de cet article est de montrer que les structures conceptuelles construites à partir de textes ne sont pas des ontologies au sens d'une conceptualisation d'un domaine au-delà de tout discours, mais qu'elles relèvent de la sémantique lexicale : *il n'y a pas de concepts dans un texte, mais uniquement des usages linguistiques de ces concepts*. La construction d'ontologies à partir de textes repose sur un ensemble d'hypothèses fortes. Avant de les considérer, rappelons ce que nous entendons par *ontologie*.

### 4.1   Définition de l'ontologie

Il existe aujourd'hui, dans le domaine de l'informatique et de l'ingénierie des connaissances, un certain consensus sur la définition d'une ontologie, consensus autour d'une définition *à la* Gruber (Gruber 1992) que nous pouvons résumer en disant qu'*une ontologie est une conceptualisation d'un domaine partagée par une communauté d'acteurs. C'est-à-dire un ensemble de concepts et de relations définis à l'aide d'un langage formel compréhensible par un ordinateur*. A cela il faut ajouter une autre définition qui, puisque les ontologies ont parmi leurs objectifs de faciliter la communication entre acteurs qu'ils soient humains ou logiciels, considère *une ontologie comme un vocabulaire de termes dont la signification est définie de manière formelle*.

Il est important de souligner les visées normatives de ces deux définitions et de rappeler les propriétés attendues, à savoir qu'une ontologie se doit d'être consensuelle – on ne peut communiquer que si l'on s'accorde sur les termes utilisés – et cohérentes, et donc partageables et réutilisables. Ces propriétés trouvent leurs origines dans l'étymologie du mot (nous y reviendrons dans le chapitre 6) : *une ontologie n'est pas une base de connaissances comme une autre*.

### 4.2   Les hypothèses du processus de la construction à partir de textes

La construction d'ontologies à partir de textes repose sur un ensemble d'hypothèses fortes. La première est de dire que les experts peuvent traduire leurs connaissances ontologiques du domaine dans des textes et que ces derniers constituent un monde plus ou moins clos. La deuxième considère le processus de rétro-ingénierie comme possible, basée sur le fait que certaines catégories de mots et que certaines relations linguistiques traduisent un usage en langue de concepts et de relations conceptuelles. La troisième hypothèse postule que les structures lexicale et conceptuelle sont relativement isomorphes. Enfin, que la validation par les experts suffit à ériger la structure conceptuelle comme une ontologie du domaine.

### 4.3 Traduttore, Traditore[11]

Ces hypothèses sont séduisantes. Les textes et les mots qui les composent, ainsi que les relations linguistiques qui les lient, sont des données objectives sur lesquelles nous pouvons appliquer des méthodes mathématiques. La démarche est scientifique. Mais, même si les informations que nous pouvons extraire des textes sont intéressantes et utiles – la structure lexicale entretient bien un rapport avec la structure conceptuelle du domaine – force est de constater que les résultats ne répondent pas à nos attentes, même lorsque nous utilisons des langages contrôlés. Qu'en est-il du consensus, de la cohérence, du partage et de la réutilisabilité des ontologies ainsi construites ? Nous pourrions bien sûr évacuer les propriétés que nous n'arrivons à pas à vérifier et changer la définition de l'ontologie en disant qu'elle reflète une des façons dont la connaissance peut être perçue et dite (Aussenac-Gilles 2005).

Il serait peut-être préférable de dire que la construction d'ontologies peut s'appuyer avec profit sur l'extraction de connaissances à partir de textes, mais que la structure conceptuelle construite à partir de corpus n'est pas une ontologie. La conceptualisation du domaine ne se calque pas sur la structure lexicale : "*le lexique des langues ne reflète pas la conception scientifique du monde*" (Rastier 2004). Cette structure relève de la sémantique lexicale et plus généralement de la linguistique textuelle dont un des principes est *l'incomplétude des textes*. Cela signifie que la compréhension des textes, que le sens[12] des mots, requiert des connaissances extralinguistiques qui par définition ne sont pas incluses dans les textes – il suffit de penser à la sémantique du nom propre non plus comme *désignateur rigide* mais comme désignant un ensemble de propriétés (*C'est un Napoléon israélien.*) pour s'en convaincre –. Et que dire de l'intention de l'auteur ? "*the meaning (in general) of a sign needs to be explained in terms of what users of the sign do (or should) mean*" (Grice 1957)[13].

Reprenons notre application sur les relais et considérons la structure lexicale (fig. 2) et l'ontologie construite par les experts à l'aide d'un langage formel (fig. 4).

---

[11] "Traduire, c'est trahir"

[12] Nous distinguons *sens* et *signification*. Ici, le sens est une signification contextualisée qui se construit en discours (et non que la signification est un sens normé).

[13] Ce qui permet d'expliquer la présence au sein d'un même corpus (dans le cadre d'une autre application réalisée pour EDF R&D) de phrases qui semblent se contredire : « *une turbine Kaplan est une turbine à hélices* », « *une turbine à hélices est une turbine Kaplan* », « *une turbine Kaplan ressemble à une turbine à hélices* », etc.

L'expression « relais de tension » est en fait un raccourci pour l'expression plus *complète* « relais à seuil dont la valeur seuillée est la tension », plus complète au sens où elle désigne de manière précise l'objet référencé. L'expression « relais de tension » dénote le concept <relais à seuil de tension> de l'ontologie formelle dont le nom peut être normalisé en « relais à seuil de tension » sans être pour autant un terme d'usage.

Cet exemple est une illustration de l'utilisation courante de figures de rhétorique, telle que la métonymie[14], l'ellipse[15] ou la synecdoque, dans l'écriture de textes techniques. Ainsi la synecdoque, qui consiste à utiliser un syntagme à la place d'un autre lorsque le premier est lié au second dans un rapport logique (partie-tout, singulier-général, etc.), est une pratique courante d'économie de mots tout en gardant une précision dans la référence. Mais *l'utilisation de tropes suppose que l'auteur et le lecteur partagent la (une) même et préexistante conceptualisation du monde nécessaire à la compréhension de ces expressions*. Cette conceptualisation, qui est en fait l'ontologie du domaine, n'est pas incluse dans les textes. C'est en référence à cette ontologie que les experts valideront la structure conceptuelle déduite de la structure lexicale, tout comme ils justifient l'utilisation des figures de style. Un certain nombre de problèmes peuvent être évités si l'on garde en mémoire que si les notions de *langue de spécialité* (ou *langues spécialisées*), *terminologie* et *ontologie* entretiennent certains rapports, elles ne se recouvrent pas (Roche 2005). Ainsi, si la langue s'intéresse prioritairement aux relations entre signifiants et signifiés, la terminologie et l'ontologie s'intéressent principalement aux rapports entre concepts et objets : *le signifié n'est pas un concept*.

## 5 Langages formels

> *« Le meilleur moyen pour éviter la confusion des mots qui se rencontrent dans les langues ordinaires, est de faire une nouvelle langue, & de nouveaux mots qui ne soient attachés qu'aux idées que nous voulons qu'ils représentent »*
> La logique ou l'art de penser. Arnauld et Nicole, Chap. XII

*Une ontologie*[16] *relève d'une conception scientifique du monde*. Elle a pour principal objectif la définition, dans un langage compréhensible par un ordinateur, des concepts nécessaires à l'appréhension et à la représentation des objets qui peuplent notre réalité de telle sorte que cette définition soit consensuelle, cohérente, partageable et réutilisable.

La construction d'ontologies par les experts à l'aide d'un langage formel doit permettre d'atteindre un tel objectif. En effet, un des premiers avantages d'une approche scientifique n'est-il pas celui de « sortir » de la langue usuelle afin de

---

[14] Trope fondé sur un rapport d'équivalence entre des termes.
[15] Emploi d'un syntagme considéré comme tronqué par rapport à une forme dite normale : « relais de tension » pour « relais à seuil de tension ».
[16] Que cette ontologie ait une visée universelle, restreinte à un domaine d'applications ou limitée à un type de tâches.

*Dire n'est pas concevoir*

modéliser le monde selon une théorie formelle, axiomatique ? Le système formel se suffit à lui-même et ne nécessite aucune connaissance qui lui soit extérieure. Nous bénéficions de tous les avantages des méthodes hypothético-déductives : si nous acceptons les postulats du système nous sommes obligés d'en accepter *ipso facto* leurs constructions, ici la définition des concepts. Ceci explique pourquoi aujourd'hui les langages à base de logique, telles que la logique des descriptions (Baader *et al.* 2003), connaissent un tel succès. La syntaxe et la sémantique sont claires et la logique nous assure un système cohérent. Mais quel est le prix à payer pour cela ? Et sommes-nous assurer des propriétés de consensus, de partage et de réutilisabilité ?

L'hypothèse de Sapir-Whorf (Whorf 1956) sur l'interdépendance de la langue et de la pensée s'applique également aux langages formels. Cela signifie que la manière dont l'ontologie est construite et le résultat obtenu (les définitions des concepts et de leurs relations) dépendent directement du langage qui est utilisé et des principes épistémologiques qui le sous-tendent. Ainsi, pour les langages qui se fondent sur la logique du premier ordre, un concept, dans la filiation de Frege, est une fonction à valeur prédicative. Le concept est clairement défini : c'est une formule bien formée. Mais alors quelle est la nature des objets subsumés par un concept défini comme la négation d'une formule ? Comment distinguer un prédicat unaire décrivant un concept d'un prédicat unaire représentant une propriété ? Comment distinguer les relations binaires décrivant des relations (rôles) de ceux décrivant des attributs ? Et enfin comment distinguer les ensembles des concepts[17] ? La logique est un langage épistémologiquement neutre, ce qui en fait sa force d'un point de vue formel, sa faiblesse d'un point de vue épistémologique. Certains travaux sont menés afin d'introduire des notions d'ordre épistémologique, telle que la notion de *rigidité* (Guarino *et al.* 94), (Kaplan 01). Mais ces notions n'introduisent pas de nouveaux principes sur lesquels se fonder pour la construction des ontologies, elles contraignent les valeurs des formules des mondes possibles (logique du *nécessaire*). Enfin, l'utilisation d'un même formalisme logique à la syntaxe et à la sémantique claires comme OWL[18] ou KIF[19], s'il peut nous garantir une certaine cohérence, ne garantit par pour autant le consensus, ni le partage ou la réutilisabilité de tout ou une partie de l'ontologie. Comment par exemple réutiliser des parties de TOVE (Fox 1992) ou de Enterprise Ontology (Uschold *et al.* 1997) bien que ces ontologies utilisent le même langage (KIF) et portent sur le même sujet (la modélisation d'entreprise) ? Malgré cela, terminons ce chapitre en insistant sur le fait que la logique est nécessaire, c'est sur elle que repose la cohérence de l'ontologie.

---

[17] On peut considérer d'un point de vue logique que le concept est la définition intensionnelle d'un ensemble. Mais on peut également considérer que d'un point de vue épistémologique concept et ensemble sont deux notions différentes : un concept a bien une sémantique ensembliste par les objets qu'il subsume (objets de même nature), alors qu'à tout ensemble ne correspond pas nécessairement un concept (ensemble d'objets pouvant être de nature différente dont l'état (défini par les valeurs de leurs attributs) vérifie une même propriété, celle-ci correspondant à la définition intensionnelle de l'ensemble).

[18] OWL Web Ontology Language: http://www.w3.org/TR/owl-features/

[19] Genesereth M.R. and Fikes R.E., *Knowledge Interchange Format Version 3.0, Reference Manual*, Report Logic 92-1, Computer Science Department, Stanford University, June 1992

Les langages logiques ne sont pas les seuls langages formels pour la construction d'ontologies. Les systèmes à base de schémas (Wright *et al.* 1984), plus justement qualifiés de semi-formels, reposent sur une modélisation plus *naturelle*. Les classes sont définies par un ensemble d'attributs communs à leurs instances et se structurent principalement selon une relation de *généralisation-spécialisation* (sur la base de factorisation d'attributs) et une relation *partitive*[20]. Il existe également d'autres formalismes basés sur des notions différentes pour la définition des concepts, comme la notion de différenciation (Roche 2001), (Bachimont *et al.* 2002).

## 6   Le modèle OK[21]

La construction d'ontologie est un problème d'ordre épistémologique qui requiert une démarche et un langage spécifiques. L'étymologie ayant des vertus prophylactiques, rappelons que l'ontologie est avant tout la « science de l'être en tant qu'être indépendamment de ses déterminations particulières ». C'est-à-dire la recherche d'une description *stable* de l'objet au-delà de ses déterminations particulières. Le modèle OK, pour Ontological Knowledge (Roche 2001), s'inscrit dans ce programme d'inspiration aristotélicienne. Programme néanmoins revisité dans la mesure où nous ne recherchons pas les caractéristiques intrinsèques à l'objet indépendamment de tout observateur, mais les caractéristiques qui font ce que l'objet *est* pour une communauté de pratique par delà la pluralité de ses déterminations : *l'ontologie a pour objectif une modélisation d'une intersubjectivité*. Ainsi, les attributs et leur factorisation ne constituent plus l'ossature sur laquelle se construisent et s'organisent les concepts. Il nous faut rechercher les caractères *eidétiques*[22] qui fondent la nature, l'essence, du concept. La définition du concept par *différenciation spécifique* – définition en genre et espèce – est l'exemple type de cette approche.

Tout comme certaines théories linguistiques considèrent que la langue est un système et que la langue est différence, le modèle OK postule qu'une ontologie est un système de concepts et qu'une ontologie est différence. Les concepts se définissent par différenciation spécifique. C'est-à-dire qu'un concept est défini à partir d'un concept existant en indiquant ce qui le différencie. Une telle définition permet de déduire *ipso facto* les similarités et les différences avec les autres concepts, qu'ils soient père ou frères. Les concepts se structurent alors en un arbre de Porphyre sur lequel se grefferont les attributs nécessaires à la description des instances des concepts : *un concept est plus qu'un ensemble d'attributs*. Les différences constituent donc le cœur du système. Elles définissent et différencient les concepts. Les définitions sont consistantes[23] au sens logique du terme et la recherche du consensus s'effectue non pas sur la définition des concepts mais sur l'identification des

---

[20] Relation mérologique, du grec *méros*, partie. La *Méréologie* de Stanislaw Lesniewski est un calcul des "totalités et de leurs parties".

[21] Notre objectif n'est pas ici de présenter en détail le modèle OK, ce qui aurait nécessité une comparaison avec d'autres systèmes, mais simplement d'illustrer une solution possible face aux problèmes énoncés.

[22] Un caractère est dit *eidétique* pour un objet si lorsqu'on le retranche de l'objet, celui n'est plus ce qu'il *est*.

[23] L'ensemble des différences constitue une base orthonormée.

*Dire n'est pas concevoir*

différences spécifiques sur lesquelles il est plus facile de s'accorder. Une telle approche a néanmoins un coût : la détermination des différences n'est pas chose aisée, elle requiert la présence d'experts ayant une bonne appréhension de la conceptualisation de leur domaine. A noter pour clore cette section que le modèle OK gère différentes notions de connaissances abstraites telles que le *concept* (subsumant des objets de même nature), le *concept composé*, la *classe* (regroupant des objets de même nature dont l'état[24] vérifie une même propriété) et l'*ensemble* (regroupant des objets pouvant être de nature différente dont l'état vérifie une même propriété). Enfin, les ontologies OK peuvent être traduites en OWL (Spies *et al.* 2006) et en KIF.

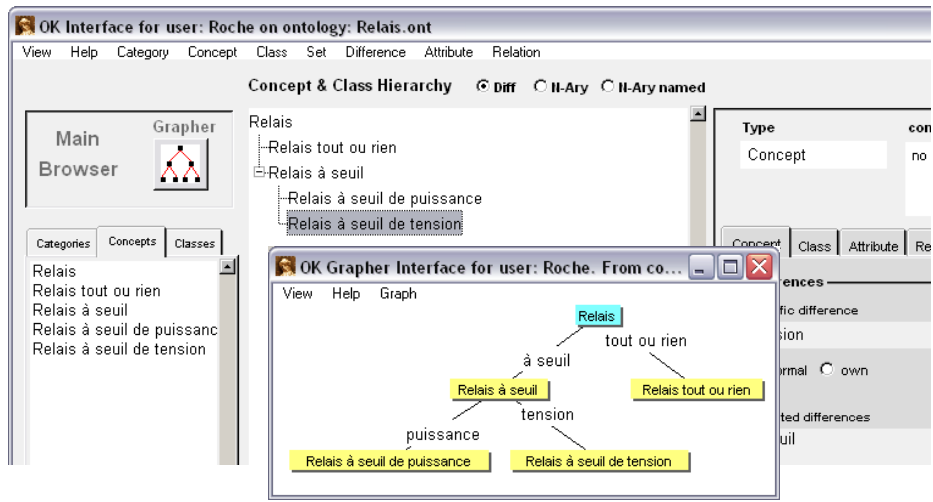

**Fig. 5** – Le modèle OK d'ontologies par différenciation spécifique (environnement OCW)

## 7   Conclusion

Nous retrouvons, dans la construction d'ontologies à partir de textes et la construction d'ontologies par les experts à l'aide d'un langage formel, les oppositions entre les approches sémasiologique et onomasiologique. Les résultats ne sont en général pas comparables. Les structures conceptuelles construites à partir de textes relèvent de la sémantique lexicale et ne sont pas à proprement parlé des ontologies au sens d'une conceptualisation d'un domaine au-delà de tout discours. Cependant l'extraction de connaissances à partir de textes est une aide précieuse à la construction d'ontologies qui reste une tâche difficile. L'extraction de candidats termes et l'identification de relations linguistiques sont autant de guides pour le choix des concepts et de leurs relations. L'utilisation de langages formels, si elle permet de garantir certaines propriétés soulève d'autres problèmes du fait de leur neutralité épistémologique. La construction d'ontologies requiert des démarches et des langages spécifiques capables de prendre en compte les différents types de connaissances nécessaires à une conceptualisation du domaine qui soit cohérente, consensuelle et partageable.

---

[24] Défini comme l'ensemble des valeurs des attributs de l'objet.

# Références